\documentclass[lnbip]{svmultln}

\usepackage{makeidx}
\usepackage{graphicx}
\usepackage{enumitem}
\usepackage{booktabs}
\usepackage{array}
\usepackage{tikz}
\usetikzlibrary{positioning,arrows.meta,shapes.geometric,fit,backgrounds,calc}
\usepackage{wrapfig}
\usepackage{hyperref}
\usepackage{tabularx}
\usepackage{longtable}
\usepackage{xcolor}
\usepackage{eso-pic}

\usepackage{hyperref}
\usepackage{graphicx}

 \newcommand{\orcidID}[1]{%
 	\href{https://orcid.org/#1}{%
 		\includegraphics[height=1.6ex]{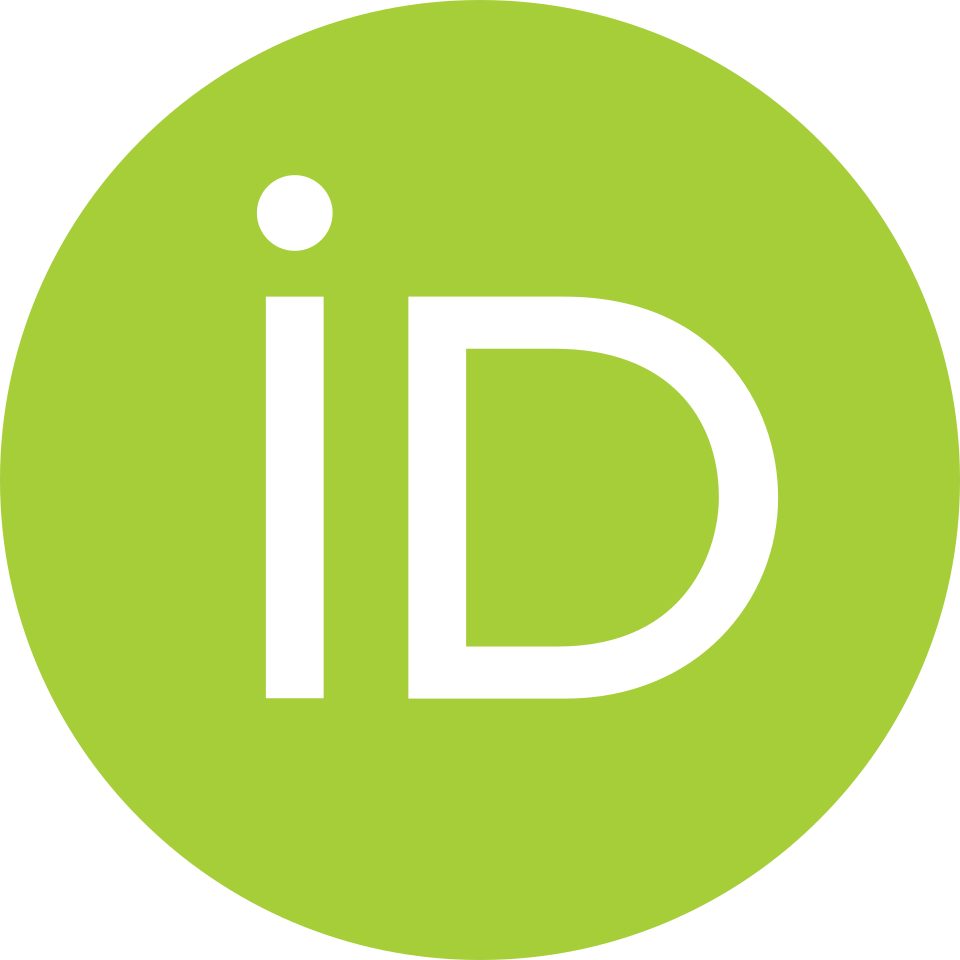}%
 	}%
 }

\begin{document}
	\mainmatter
	
	\title{AI-Decision Checkpoints for AI-Augmented Business Process Management: Framework and Educational Instantiation}
	\titlerunning{AI-Decision Checkpoints for AI-Augmented Business Process Management}
	
		\author{Amin Jalali~\orcidID{0000-0002-6633-8587}}
		\institute{Department of Computer and Systems Sciences, Stockholm University,\\ Stockholm, Sweden\\\email{aj@dsv.su.se}}

	\maketitle
	
	\begin{abstract}
		Large Language Models (LLMs), Retrieval-Augmented Generation (RAG), and AI agents are increasingly embedded in operational business processes. Yet Business Process Management (BPM) curricula and frameworks still largely treat artificial intelligence (AI) as an add-on technology, leaving graduates (as potential future process developers) unprepared to reason about AI as a first-class design element of end-to-end processes. This paper addresses that gap by proposing \emph{AI-decision checkpoints}: explicit moments in a process development trajectory where process developers identify AI-candidate sub-processes, assess expected effects on time, cost, quality, and flexibility, consider legal and organisational constraints, and document a reasoned decision to adopt, constrain, or reject specific AI components. The checkpoints are instantiated through a fictitious customer onboarding process as a \textit{BPM Teaching Case}, embedded in a lifecycle-driven framework spanning six modules that combine process modeling, simulation, workflow execution with AI agents, and process mining, with each module's output serving as the next module's input. A preliminary formative reflection draws on instructor observations, submitted artifacts, and discovered process maps from learning-management-system logs. These exploratory observations suggest that the approach supported clearer distinctions between task-level automation and process-level value.
		
		\keywords{BPM education, BPM lifecycle, curriculum design, AI-augmented BPM, teaching case}
	\end{abstract}

	\section{Introduction}
	\label{sec:intro}
	
	Business Process Management (BPM) is a mature information-systems discipline that provides concepts, notations, and methods for identifying, modeling, analyzing, redesigning, executing, and monitoring organizational processes~\cite{dumas2013fundamentals,weske2012business}. Core pedagogical content has remained remarkably stable for decades: process discovery and modeling, control-flow analysis, simulation-based performance analysis, workflow enactment, and process mining~\cite{van2016data}.
	
	This stability is now challenged by advances in Large Language Models (LLMs), Retrieval-Augmented Generation (RAG), and AI agents. It has become technically feasible to introduce AI agents as process participants in knowledge-intensive tasks~\cite{lewis2020retrieval}, such as extracting data from unstructured documents, conducting Know-Your-Customer (KYC) or fraud checks, generating customer communication, and lead generation for claim investigation~\cite{khayatbashi2025ai}.
	
	Industry adoption of such components currently outpaces their curricular integration, leaving graduates at risk of two complementary deficits. Classically trained BPM analysts can model and analyze processes, but struggle to reason about when and how AI components add value at the process level. Conversely, AI engineers may be able to build proof-of-concept agents and deploy them in production but lack the methodological grounding to assess their impact on end-to-end performance and value creation.
	
	Existing curricular responses tend to follow one of two patterns: appending a stand-alone module on AI or LLMs to an otherwise unchanged course, or adding AI-related examples throughout traditional content without altering its structure. In both cases, AI is treated as a technology topic rather than as a design concern to be evaluated systematically across the BPM lifecycle.
	
	We contend that the discipline already possesses a suitable backbone for structuring AI-related reasoning: the BPM lifecycle~\cite{dumas2013fundamentals}, which has long served as a textbook spine and didactic organizer. Our contribution is not to propose a new lifecycle, but to turn the existing one into an active driver of AI-related decision-making through what we call \emph{AI-decision checkpoints}: pre-specified points in the learning trajectory at which students must identify, justify, configure, or reject an AI component using the analytical instruments of the corresponding lifecycle phase, transforming the lifecycle from a passive organiser into a sequence of grounded decision situations. Therefore, this paper makes three contributions:
	
	\begin{enumerate}[label=(C\arabic*)]
		\item It presents a fictitious \emph{Banking case} designed to sustain multi-module engagement across the BPM lifecycle, 
		\item It introduces \emph{AI-decision checkpoints} as the organizing pedagogical artifact of the case, with one checkpoint per lifecycle phase, embedded in a lifecycle-driven curriculum framework in which each module's output becomes the next module's input across modeling, simulation, redesign, implementation, and mining,
		\item It presents a \emph{formative reflection} on a Master-level instantiation of the case, combining instructor observations with insights discovered from the learning-management-system event logs showing how groups progressed across modules.
	\end{enumerate}
	
	The remainder of this paper is structured as follows. Section~\ref{sec:related} positions the work in relation to previous related literature. Section~\ref{sec:framework} introduces the AI-decision checkpoint framework and positions each checkpoint within the BPM lifecycle. Section~\ref{sec:instantiation} describes its instantiation in a Master-level course built around the Banking teaching case. Section~\ref{sec:evaluation} reflects on course delivery and student performance based on instructor observations and learning-management-system data. Section~\ref{sec:conclusion} concludes and outlines implications for BPM educators and directions for future work.

	\section{Related Work}
	\label{sec:related}
	
	Despite BPM's maturity as a discipline, its integration into higher-education curricula has remained narrower than the scope of professional practice.
	Moormann and Bandara~\cite{moormann2012bpm} issued an early call for action, noting that graduates consistently lack the breadth of skills employers require.
	To identify related work, a Scopus search was conducted using the query shown in \tablename~\ref{fig:scopus-query}, which returned 45 documents at 2026-05-31. After title-and-abstract screening for papers addressing the intersection of BPM education and AI, three papers were identified as indirectly related. 
	
	\begin{table}[b]
		\centering
		\fbox{\parbox{0.92\textwidth}{\small
				\textbf{Scopus search query (45 results):}\\[4pt]
				\texttt{TITLE-ABS-KEY(}\\
				\texttt{\quad("business process management" OR BPM)}\\
				\texttt{\quad AND ("artificial intelligence" OR AI OR "generative AI"}\\
				\texttt{\quad\quad OR "large language model*" OR LLM*}\\
				\texttt{\quad\quad OR "AI agent*" OR "retrieval-augmented generation" OR RAG)}\\
				\texttt{\quad AND (teaching OR education OR pedagogy OR curriculum}\\
				\texttt{\quad\quad OR "course design" OR "higher education")}\\
				\texttt{)}
		}}
		\caption{Scopus query used for literature identification.}
		\label{fig:scopus-query}
	\end{table}
	
	 On curriculum innovation, Dallavalle et al.~\cite{dallavalle2025educating} propose a PBL-Process Mapping Blueprint for undergraduate BPM courses at the University of São Paulo, structured around three next-generation BPM drifts: Conversationality, Autonomization, and Sophistication. The blueprint was developed in three phases, culminating in prospective validation through two focus groups with eight alumni. Participants identified autonomization as the most professionally consequential drift and called for a brief AI-agent primer to be introduced early in the course. The blueprint treats AI as one of several emerging competencies to be woven into existing activities. Our work shares the view that AI must be embedded as a design concern throughout the lifecycle, but operationalizes it differently: rather than a primer, we provide students with a structured decision instrument at every lifecycle phase, requiring them to justify, configure, or reject AI components using phase-appropriate evidence.
	
	 On teaching cases, Grangel et al.~\cite{grangel2025teaching} present a teaching case aimed at postgraduate students in AI, BPM, or Medical Informatics, in which real pseudoanonymised clinical data from a Spanish hospital are used to analyse Failure To Rescue events through process mining and predictive process monitoring. The case covers process discovery, conformance checking, performance analysis, and process redesign, and is accompanied by online instructor-facing teaching notes. Like our Banking case, it grounds BPM competencies in a domain-specific scenario with real complexity. The key difference is scope: the Grangel et al.\ case focuses on applying mining techniques to an existing process, whereas our case requires students to design, implement, and evaluate AI-augmented process variants across the full lifecycle, treating governance concerns such as human-in-the-loop routing and fallback handling as explicit learning objectives.
	
	 On LLMs in BPM education, Kluza et al.~\cite{kluza2025revisiting} conduct two experiments in which ChatGPT-4o and Gemini Flash are tested on typical BPMN assignment tasks: answering comprehension questions based on five BPMN models (in PNG and XML format) and detecting errors in 30 flawed BPMN diagrams. The authors conclude that BPMN assignments are not inherently AI-resistant and call for AI-aware pedagogy in which students learn to critically assess AI-generated output. In our course, LLMs are not evaluated as assignment solvers but configured and deployed as process participants. The control-flow reasoning limitations that Kluza et al.\ document are precisely what motivate the AI-decision checkpoint for implementation, where students must specify schema-constrained outputs, confidence thresholds, and human escalation paths to compensate for unreliable AI behavior.

	Collectively, these works show that AI is entering BPM curricula through multiple routes, yet none provides a systematic method for reasoning about AI adoption across all lifecycle phases using the analytical instruments of each phase. The AI-decision checkpoint framework addresses this gap.
	
	\section{AI-Decision Checkpoints}
	\label{sec:framework}

	The proposed framework uses the BPM lifecycle to structure a sequence of explicit, phase-related decisions about AI adoption.
	At each checkpoint, students contribute to determining how AI can enhance processes by identifying AI-candidate subprocesses, establishing the baseline, evaluating expected benefits and risks using the analytical instruments of the corresponding lifecycle phase, and documenting a reasoned decision to adopt, adapt, postpone, or reject specific AI components. This approach decomposes AI adoption from a monolithic yes/no judgment into a chain of locally grounded assessments that distribute AI-related reasoning across all phases, anchor decisions in established BPM instruments, and produce a traceable decision trail that can be revisited.
	
\begin{figure}[t]
		\centering
		\includegraphics[width=0.8\textwidth]{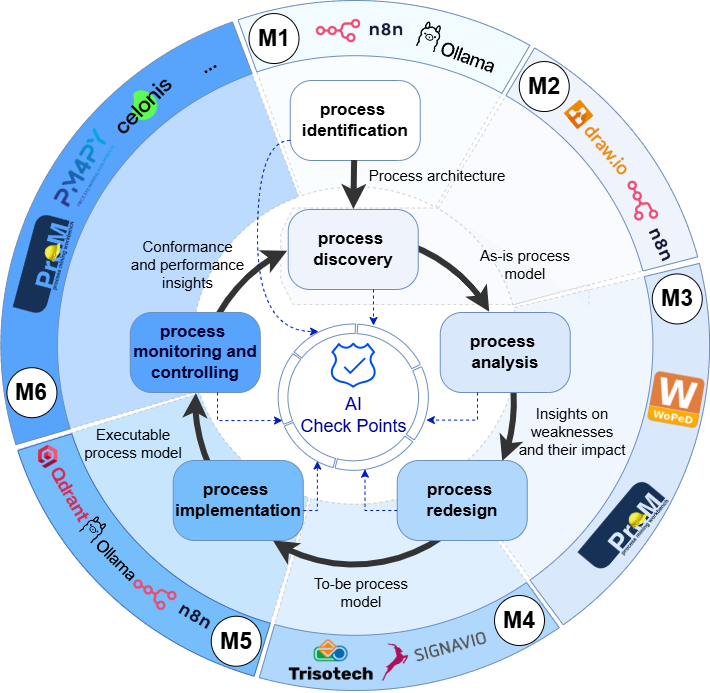}
		\caption{The BPM lifecycle with six AI-decision checkpoints and their instantiation in course modules (M1-M6) with associated tool support. We operationalized CP4 in two modules in our course: an initial simulation-informed redesign in M3 (workflow-net level) and a final exception-handling redesign in M4 (BPMN level).}
		\label{fig:lifecycle}
	\end{figure}

	The framework does not prescribe a correct answer at any checkpoint; it foregrounds reasoning quality, requiring students to articulate assumptions, acknowledge trade-offs, and justify decisions. One checkpoint aligns with each lifecycle phase: identification, discovery, analysis, redesign, implementation, and monitoring~\cite{dumas2013fundamentals}; individual checkpoints can be weighted or combined without disrupting the logic. \figurename~\ref{fig:lifecycle} positions them relative to the six course modules (M1-M6).

	\subsection{CP1 - Process Identification: Scoping AI Opportunities}
	
	CP1 is located in the process-identification phase. Its purpose is to determine in which processes and subprocesses AI augmentation is sufficiently promising to warrant further investigation. The guiding question is: \emph{which elements of the process landscape are plausible AI-candidate subprocesses, and what effects are they expected to have at the level of entire processes?}
	
	The process architecture is examined for subprocesses amenable to AI, specifically those exhibiting high knowledge intensity, data-processing demands, repetitiveness, or dependence on scarce expertise. Each candidate's expected impact is assessed along the \emph{devil's quadrangle} (time, cost, quality, flexibility)~\cite{reijers2005best}, and risks are cataloged covering legal and regulatory requirements (e.g.\ GDPR), ethical considerations, data availability, and organizational readiness. The outcome is an AI opportunity register that combines benefit assessments with initial technical feasibility in a prioritization matrix, determining which candidates proceed to later phases.

	\subsection{CP2 - Process Discovery: Positioning AI in the As-Is Process}
	
	CP2 is located in the process-discovery phase. The focus shifts from the process landscape to a concrete as-is process, and from abstract opportunity identification to a precise understanding of how candidates from CP1 relate to existing organizational work. The guiding question is: \emph{given a description of the selected process, where would the AI candidates conceptually attach, and which human and system actors would they interact with or affect?}
	
	The as-is process in this phase shall be specified as a normative model and enriched with an explicit actor perspective. Candidates prioritized in CP1 are positioned with respect to this model rather than integrated into it. For each candidate, the analysis specifies (i) the segment of the as-is model it would conceptually replace, support, or augment, (ii) the human roles and IT systems currently involved at that segment and how they would be affected, and (iii) the inputs, outputs, and control transfers currently observed. Escalation paths, approval steps, and exception-handling mechanisms that would become relevant under AI augmentation are also sketched at a conceptual level.
	
	The artifact is the as-is process model together with an AI-positioning note that records, for each candidate, its intended placement, the actors affected, and prominent structural obstacles or dependencies.
    
    \subsection{CP3 - Process Analysis: Performance Baseline}
    
	CP3 is located in the process-analysis phase. While CP1 and CP2 establish where AI augmentation is promising and how it relates to the as-is process, CP3 quantifies the current performance without altering the process structure. The guiding question is: \emph{given an executable representation of the current process, what is the baseline performance against which any redesign (AI-based or otherwise) must be assessed?}
	
	The baseline can be identified in different ways such as extracting the current state of the process using process mining, or transforming the model from CP2 into an executable model and simulating the process to identify the baseline.	
	In courses without prior process data, the latter approach can be feasible; while in practical settings, other approaches can be followed. Regardless of the method, the analysis should yield baseline indicators such as average throughput times, waiting times, resource utilization, and sensitivity to workload variation.
	
	The artifact produced at CP3 is a baseline performance report documenting the chosen method, underlying assumptions, and key performance indicators, with particular attention to structural bottlenecks and systematic delays. The importance is to define a baseline against which upcoming changes can be evaluated (whether AI-driven or otherwise). Crucially, where a simulation model is established as part of this baseline, it serves as a reusable analytical asset: as redesign proposals emerge in later phases, the model can be refined to reflect the proposed changes and re-simulated to assess their expected impact before any real implementation takes place.

	\subsection{CP4 - Process Redesign: Value-Aware Business Process Redesign}
	
	CP4 is located in the process-redesign phase. Building on the baseline from CP3 and the conceptual positioning from CP1 and CP2, it treats AI augmentation as one design option within a broader redesign effort. The guiding question is: \emph{when redesign practices are applied, under which conditions does the inclusion of AI-candidate subprocesses produce to-be process models that outperform the baseline?}
	
	Starting from the executable as-is model and diagnosed weaknesses, the redesign uses recognized principles (task elimination, resequencing, parallelization, integration of responsibilities). To-be scenarios include both classical control-flow and resource changes and variants that embed AI candidates, integrating AI with redesign moves rather than bolting it on. Redesign variants are distinguished by the AI technology embedded: (i)~LLM-based agents for email intake and structured data extraction from documents, (ii)~RAG-equipped agents for fraud screening that retrieve and reason over AML rule documents, and (iii)~classical BPM redesign variants without AI components, which serve as the comparison baseline. All scenarios are simulated under consistent assumptions and compared to the baseline.
	
	The artifact is a comparative redesign-and-simulation dossier in which each candidate is assigned a redesign decision, justified by the redesign rationale and quantitative evidence, including trade-offs such as bottleneck relocation or increased process complexity.
	
	\subsection{CP5 - Process Implementation: Realisability and Governance}
	
	CP5 is located in the implementation phase and addresses the gap between conceptual redesign and technical realization. The guiding question is: \emph{are the AI-augmented redesign decisions technically realizable within the selected workflow and AI infrastructure?}
	
	The to-be design is mapped onto a concrete workflow platform and integrated with specific AI services, such as LLM APIs or RAG pipelines. For each AI component, required data flows, performance characteristics, error-handling mechanisms, and security and privacy constraints are matched against what the infrastructure can provide. Where the platform does not natively support the required control-flow patterns or error-handling logic, compensating structures are introduced, for instance, additional validation activities, schema-constrained prompts, explicit human approval tasks, or fallback paths triggered on AI failure or low confidence.
	
	The artifact is an implementation feasibility and governance dossier per AI component. It records the extent to which the component could be realized as designed, the workarounds introduced, the governance measures put in place (logging, thresholds, escalation rules, human-in-the-loop points), and any residual technical or organizational risks.

	\subsection{CP6 - Process Monitoring: Mining-Based Insights}
	
	CP6 is located in the monitoring-and-controlling phase and concerns the ex-post evaluation of AI-augmented subprocesses on the basis of execution data. The guiding question is: \emph{given execution data of the deployed process, do the AI-augmented subprocesses behave as intended and deliver net value over time, or should the process be redesigned and reconfigured?}
	
	Execution data are collected as event logs once the deployed process has run sufficiently; in a course setting, pre-generated logs are typically provided. Process discovery is applied and tested against the normative to-be model from CP4/CP5; conformance analysis surfaces deviations such as bypassed AI-enabled subprocesses; and performance analysis quantifies indicators against the CP3 baseline and CP4 simulated expectations. The artifact is a mining-based evaluation summary consolidating execution patterns, conformance deviations, and realized performance. Its insights feed back into CP1, completing the evidence-based governance cycle.
	
	\section{Course Instantiation}
	\label{sec:instantiation}

	The framework was instantiated in a Master 's-level course for students in the final year of a Business Information Systems or Computer Science program. 
	The course has been coordinated by the author for more than 15 years and is designed based on experience-based learning~\cite{jalali2018teaching}.
	No prior knowledge of BPM or AI was assumed. A total of 152 students enrolled in the course. Of these, 29 had passed the group assignments in a previous year and were required to participate only in the individual exam; the remaining 123 were new students who formed 25 project groups, each with a maximum of five members. The course mandates group work, with results delivered at each module milestone. A key structural principle is the \emph{evolving-artifact} organization: each module's output becomes the next module's input, so students revisit AI-related decisions repeatedly with accumulating evidence. 
	
	\subsection{The Banking Case}
	\label{sec:case}
	
	The primary deliverable of this paper is the Banking case. The six modules (M1-M6) are built around this case, where students focus on designing a customer onboarding process comprising customer registration, fraud assessment, and approval, at a fictitious retail bank facing a scalability crisis. The case is intentionally rich in regulatory constraints, knowledge-intensive work, and heterogeneous data sources, which together make it suitable for realistic AI augmentation scenarios. It is structured around two main documents alongside other instructions distributed to students at different points in the course.
	
	\emph{Document A} (released at M1, needed for M2) describes the current as-is process. Seven roles are involved: a clerk who registers scanned customer requests, a credit investigator who checks customer credits via an external service, two CRM officers who manage customer communications and profile activation, an accounting assistant who enables transactions, and two fraud investigators who manually review each case against the bank's anti-money-laundering (AML) guidebooks. The normal scenario traces a case from registration through credit check, parallel profile and account activation, customer notification, fraud review, and either full access or external fraud investigation. 
	A cancellation scenario (needed and can be released at M4) introduces a regulatory constraint on when a customer can withdraw the case, which is before an external fraud investigation returns a positive result, requiring a carefully bounded cancel region.
	
	\emph{Document B} (released at M3) describes the redesign brief: the bank cannot scale with the current manual process and proposes a combination of AI agents, where students are guided to prioritize email-based registration extraction and RAG-based fraud screening, as well as web-service automation (credit checks, profile and transaction management, police reporting). The fraud-screening agent is deliberately constrained by a known class-imbalance problem (the data science team reports high recall but imperfect precision), so students must reason about when the agent's output can be trusted autonomously and when a human must remain in the loop.
	
	The hook of the case is established in M1: students build a working prototype of the email-intake AI agent before they have modeled the process, then are asked whether deploying it would actually help the bank. The tension between ``the agent works'' and ``we don't yet know if it adds value'' motivates the entire lifecycle journey that follows. Case materials, module instructions, and datasets are openly available in a GitHub repository~\footnote{\url{https://github.com/jalaliamin/AI-decision-checkpoints/tree/main}}. Teaching notes, discussion guidance, and sample solutions are available to instructors upon request.

	\subsection{Module 1 - Process Identification, BPM and AI Foundations, and CP1}
	The first module (M1) introduces the conceptual foundations required to apply the framework: a concise overview of BPM and its evolution, the BPM lifecycle, and core notions such as processes, subprocesses, and actors, together with an introduction to the AI technologies most relevant to BPM: Large Language Models (LLMs), generative AI, and AI agents.
	
	On this basis, the banking case is presented, and students construct a process architecture for the bank. Within this architecture, they identify AI-candidate subprocesses, documenting their assumptions as artifacts introduced in CP1. In the onboarding context, two prominent candidates are expected to emerge from the given case description: a subprocess handling incoming customer emails and registration requests (including data extraction from attached documents), and a subprocess for assessing potential fraud against the bank's internal guidelines and regulatory documents.
	
	A complementary lab makes the building blocks tangible: students configure an AI agent in \texttt{n8n} that processes customer emails, extracts information from attached PDF forms using a local LLM (served by \texttt{Ollama} to meet on-premises regulatory constraints), and registers the result into a mock system. The lab surfaces the distinction between task-level automation and process-level value.
	
	CP1 is discharged by applying the AI opportunity register to the two candidates, yielding a prioritization matrix that determines which proceed to M2.

	\subsection{Module 2 - Normative As-Is Modeling and CP2}
	
	M2 focuses on normative as-is modeling in \texttt{draw.io} using workflow and presentation patterns. CP2 is discharged by positioning the CP1 AI candidates (email-intake and fraud assessment) with respect to the as-is model. Students also evaluate \texttt{n8n}'s workflow-pattern support from a chief-IT-officer perspective, producing the AI-positioning note that feeds into M3.

	\subsection{Module 3 - Process Analysis and Redesign, and CP3 and CP4}

	M3 covers process analysis and an initial redesign pass. Students transform the as-is model into a workflow net (\texttt{WoPeD}), verify soundness, configure simulation parameters, and obtain baseline indicators; performance spectrum analysis in \texttt{ProM}~\cite{denisov2018performance} surfaces bottlenecks. Building on this baseline, students construct and simulate redesign variants (from classical changes such as task elimination, resequencing, and parallelisation~\cite{reijers2005best} to AI-embedding designs), producing a preliminary CP4 report to be finalized in M4. Splitting CP4 across M3 and M4 is intentional: the workflow-net redesign grounds design choices analytically before the BPMN exception-handling redesign in M4.
	
	\subsection{Module 4 - BPMN-Centric Process Redesign and CP4}
	
	M4 refines the best M3 scenario in BPMN (using \texttt{Signavio} and \texttt{Trisotech}), shifting from happy-path to exception-handling redesign. Students apply established modeling guidelines to clarify control flow, make subprocess boundaries explicit, and model human-system-AI interactions and cancellation scenarios. This enables a second pass on CP4, revisiting AI-augmented subprocess choices in light of representational clarity and implementability.
	
	The artifact is a consolidated BPMN-based redesign model with clearly identified AI-augmented subprocesses and associated redesign principles. Together with the analytical evidence from M3, this model constitutes the final outcome of CP4 and provides the basis for the feasibility and governance analysis in M5.
	
	\subsection{Module 5 - Implementation in Workflow Platforms and CP5}

	The fifth module (M5) addresses the implementation phase and operationalizes CP5. Building on the selected and refined BPMN-based to-be model from M4, students map the preferred redesign onto a concrete AI-augmented workflow in \texttt{n8n}. The emphasis is on realizing the control-flow structures and subprocess boundaries defined in BPMN and on technically integrating the AI-augmented subprocesses retained after CP4.
	
	Students decompose the BPMN model into an executable workflow and realize AI-augmented subprocesses as concrete agents: an email-intake agent for document extraction and a RAG-based fraud-assessment agent that retrieves and reasons over AML rules embedded in \texttt{Qdrant}, with a human-in-the-loop step for flagged cases. The fraud agent introduces students to configuring semantic memory within a workflow platform.

	CP5 is discharged by assessing, for each AI-augmented subprocess, the extent to which the BPMN-level design from M4 can be realized and governed in the chosen technical environment. 
	The resulting dossier documents (i) alignment between design and implementation, (ii) workarounds and additional control structures introduced (validation and enrichment steps, schema-constrained prompts, fallback paths on AI failure), and (iii) governance and oversight mechanisms. On this basis, the redesign-level decisions from CP4 are refined into implementation decisions.
	
	\subsection{Module 6 - Process Mining and CP6}
	
	M6 instantiates CP6 via process mining. Students receive a multi-year synthetic event log (constructed to exhibit concept drift). They segment the log, characterize drifts, discover period-specific models, and evaluate them on fitness, precision, simplicity, and generalization. The artifact is a mining-based evaluation summary consolidating execution patterns, drift evidence, conformance deviations, and performance indicators that feed back into CP1.

	\section{Reflection on Course Delivery and Student Performance}
	\label{sec:evaluation}
	
	The course's core has been offered by the instructor since 2011; its core scaffolding was therefore pedagogically validated prior to the present intervention, which restructured it around the AI-decision checkpoints and adopted \texttt{n8n} as the implementation platform, representing a substantial redesign effort of approximately 3 months. The course follows a project-based learning (PBL) design~\cite{gdowska2025project}. At the time of writing, the exam has not yet been administered, so the evaluation focuses on group performance in labs and assignments.
	
	Given the single-course, single-instructor design with no comparison group and no final exam data at the time of writing, all findings below should be read as a preliminary, exploratory reflection on experience rather than as evidence of learning effectiveness. The evaluation addresses three questions: (EQ1) Do students engage with AI adoption as expected phase-specific decision? (EQ2) How is submission timing distributed across modules, and what does this imply for course design? (EQ3) How do students perceive the learning value of each module's primary activity? 
	Evidence triangulates instructor observations with insights discovered from LMS event logs extracted by PM4Moodle~\cite{miri2026object}, a voluntary end-of-course group survey (conducted per groups), which capture group interactions at checkpoints. Given the single-instantiation design, findings are presented as preliminary indications rather than generalizable results.

\begin{figure}[t]
		\centering
		\includegraphics[width=\linewidth]{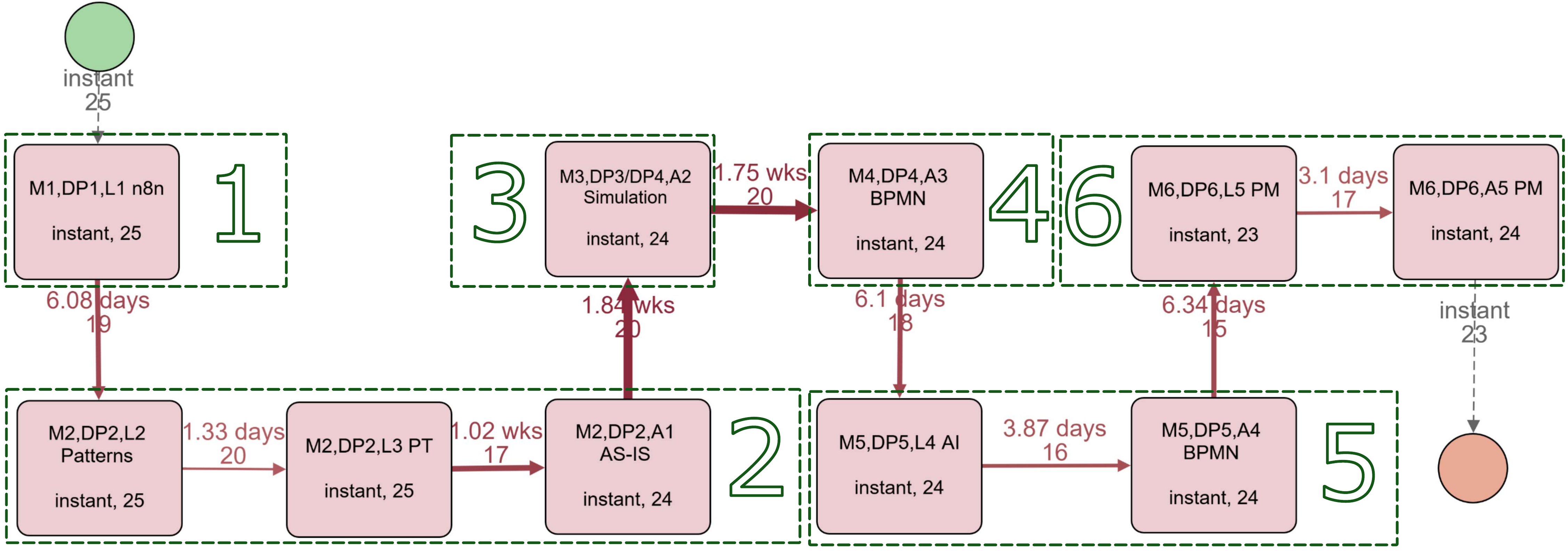}
		\caption{Discovered process map of student-group progression through labs and milestones, annotated with activity frequencies and throughput times.}
		\label{fig:process-map}
	\end{figure}

	\figurename~\ref{fig:process-map} shows a process map discovered with Apromore, using each group as a case identifier and retaining only the main paths; activity durations are instantaneous as the learning management system only captures the completion timestamps for submission boxes, but module durations can be estimated by the sum of actions taken after the previous module until the last touchpoint in each module. Several findings emerge.

	\textbf{\emph{Groups and progression paths:}} Of 25 registered groups, 24 were active at the time of analysis; one discontinued after M1. All 24 continuing groups followed the expected sequential progression, suggesting that the checkpoint artifacts were sufficient to proceed. 
	
	\textbf{\emph{Submission timing and effort distribution:}} Because the LMS captures only submission timestamps, module durations are approximated as the interval from the previous module's last touchpoint to the current module's submission. With this caveat, groups appeared to spent between 9 days and 2 weeks on each module after M1. This pattern is consistent with student feedback, suggesting that certain modules should be lightened to reduce overall course density.
	
	\textbf{\emph{Evidence on EQ1 (revised AI assessments):}} Checkpoint artifacts were reviewed by the instructor against criteria assessing (i)~completeness of the AI opportunity register or redesign dossier, (ii)~quality of justification for AI adoption decisions, and (iii)~consistency of reasoning with evidence accumulated in prior modules. This review was qualitative and based on instructor judgment; detailed rubrics will be shared in the public repository. To illustrate how checkpoints encouraged students to evaluate AI adoption during process development, we describe a pattern observed in the CP4 artifacts. Most groups identified the email-intake and fraud-assessment subprocesses as AI candidates in CP1 after receiving feedback, and the majority initially expected substantial reductions in throughput time; in several cases, they estimated that AI-based intake would halve overall case cycle time. After completing the baseline simulation in CP3, these groups found that the bottleneck lay not in registration but in the two-person fraud-investigation queue, which operated near full utilization. 	
	Their CP4 comparative redesign dossiers explicitly recorded revised assessments: the email-intake agent, while technically functional, was projected to shift the bottleneck downstream rather than reduce end-to-end cycle time unless the fraud-investigation subprocess was redesigned with the help of a second AI agent equipped with RAG. These groups consequently reclassified the intake agent from a high-priority candidate to a medium-priority one, conditional on redesigning the fraud investigation with the help of an AI agent and applying a re-sequence pattern to do it earlier in the process. This pattern of initial overestimation followed by evidence-driven revision was evident in the written justifications attached to CP4 artifacts and was confirmed through instructor observations during group supervision sessions. It illustrates the intended function of the checkpoint chain: not to discourage AI adoption, but to ground it in process-level evidence accumulated across phases. A comparable pattern of evidence-driven revision appeared at CP5: several groups that had modeled a fraud-screening path in BPMN (M4) introduced an explicit human-approval step in their \texttt{n8n} implementation once they encountered the structured-output reliability limitations of the local LLM, converting a conceptual governance requirement into a concrete workflow control.
	
	\textbf{\emph{Process identification phase (M1):}} No dedicated submission point was designed for students to submit their process architecture and deliverables related to CP1, as communication for this phase occurred primarily through in-class presentations. The absence of such a point limits data capture, and we therefore plan to introduce a dedicated submission box in the next course offering.

	\textbf{\emph{Process discovery phase (M2):}} Took longer than expected; the cancel-region pattern proved demanding and will be deferred to M4 in the next offering. Most groups also postponed formal submissions until the deadline despite completing labs early; future offerings will require submission immediately after group presentations.
	
	\textbf{\emph{Process analysis and redesign (M3):}} Took 1.8 weeks, in line with expectations. Applying redesign patterns consumed most effort; re-simulation after model changes proceeded quickly. \texttt{ProM} performance spectrum analysis required more time than \texttt{WoPeD}'s built-in KPIs.
	
	\textbf{\emph{Process redesign using BPMN (M4):}} Took 1.7 weeks, as expected. Students needed to recap cancellation patterns introduced in M2; deferring those to M4 (planned for next offering) should reduce this overhead. Despite this, students engaged more purposefully with exception-handling patterns once their practical role in redesign was apparent.
	
	\textbf{\emph{Implementation phase (M5):}} Students completed M5 faster than the expected two weeks, attributed to cumulative scaffolding: they had already built an AI agent (M1), evaluated \texttt{n8n}'s pattern support (M2), and modeled agent roles and exception paths (M3--M4). Prior checkpoints appeared to pay off at implementation time.
	
	\textbf{\emph{Process monitoring and mining phase (M6):}} This module proceeded as expected; on average, students submitted both the assignment and the lab within 9 days.
	
\begin{figure}[t]
		\centering
		\includegraphics[width=\linewidth]{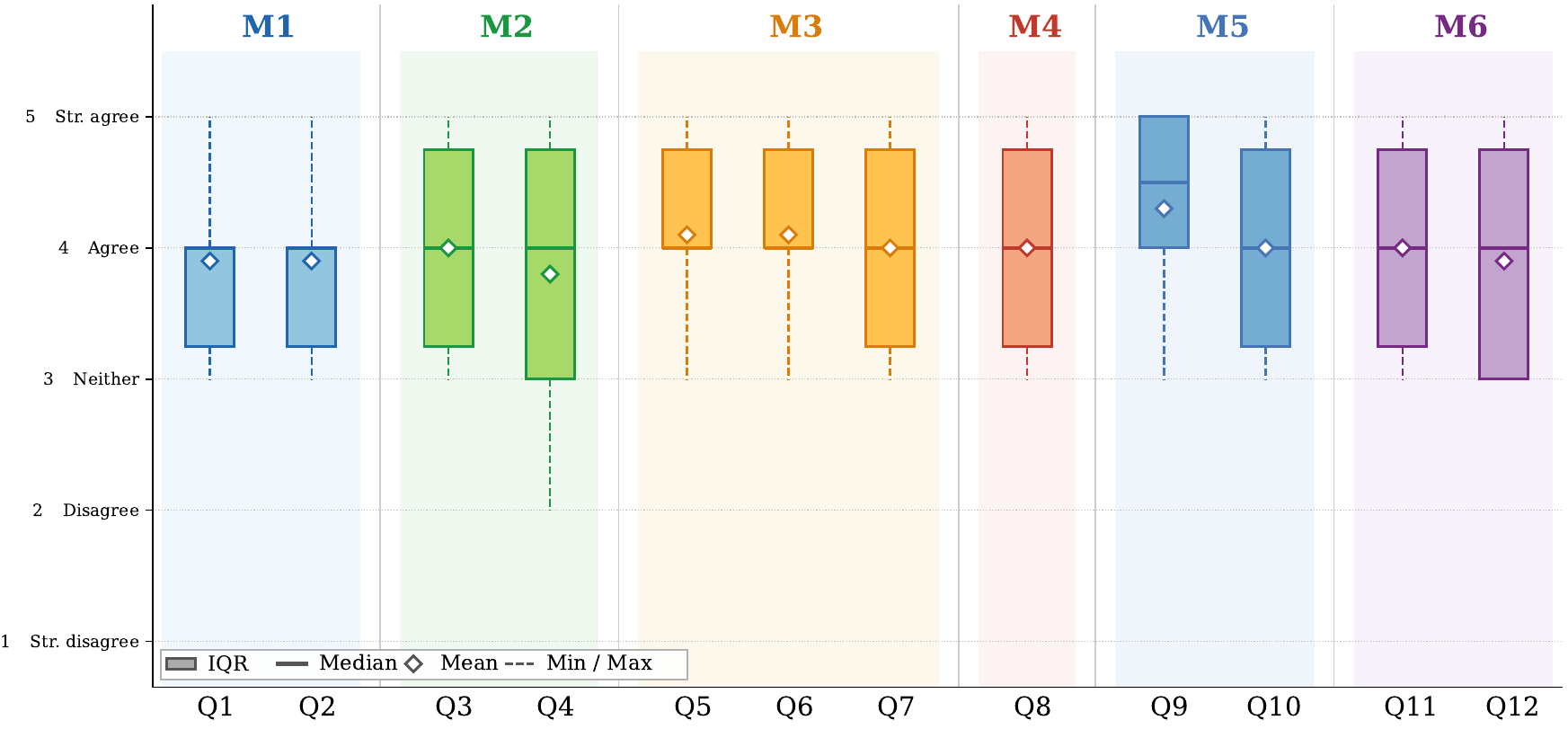}
		\caption{Perceived learning value per module activity (10 group responses, 5-point Likert scale). Boxes show the IQR; horizontal lines indicate medians; diamonds indicate means; dashed whiskers extend to min/max.}
		\label{figure:survey}
	\end{figure}

	\textbf{\emph{Evidence on EQ3 (student perceptions):}} A voluntary end-of-course survey (10 of 24 groups responded) had each group rate 12 Likert statements (1--5) on the perceived learning value of each module's primary activity: Q1--Q2 (M1), Q3--Q4 (M2), Q5--Q7 (M3), Q8 (M4), Q9--Q10 (M5), Q11--Q12 (M6). \figurename~\ref{figure:survey} summarizes the responses; full survey items are available in the public repository~\footnote{\url{https://github.com/jalaliamin/AI-decision-checkpoints/blob/main/Survey_Questions.md}}.
	
	Overall ratings are positive: all 12 items have a median of 4 (Agree), and no item falls below 3. The M5 implementation items show the highest ratings (Q9: mean\,=\,4.3, median\,=\,4.5), consistent with the LMS-based analysis finding that students completed M5 faster than expected, corroborating the interpretation that earlier checkpoints provided effective scaffolding for implementation. Q4 (workflow-pattern analysis in \texttt{n8n}) shows the widest spread (min\,=\,2, mean\,=\,3.8), reflecting variation in groups' prior experience with workflow engines. The survey is limited by a 42\,\% group response rate and single-instantiation scope; results should be treated as indicative rather than conclusive.

	\subsection{Threats to Validity and Limitations}
	\label{sec:threats}
	
	Several limitations bound the interpretation of the findings above and warrant explicit discussion.
	
	\textbf{\emph{Internal validity:}} The reflections rely on a single instantiation of the framework delivered by the authors. Researcher-instructor overlap may have shaped both the design and the interpretation of student behavior, even though triangulation with LMS event logs partially mitigates this risk. Furthermore, the course is still in progress, and individual learning outcomes via the final exam have not yet been assessed.
	
	\textbf{\emph{Construct validity:}} The LMS captures only those interactions that occur at the predefined touchpoints. Cognitive work conducted outside the LMS (peer discussion, off-platform prototyping, or iterative refinement of models) is not directly observable and is approximated indirectly through artifact submissions and teaching-team observations~\cite{karunaratne2025and}. The artifacts produced at each checkpoint also vary in granularity, which constrains comparability across phases.
	
	\textbf{\emph{External validity:}} The framework was instantiated in a single Master-level program with a banking onboarding case. While the case is intentionally rich in regulatory constraints and heterogeneous data, generalization to other domains, study levels, or institutional contexts requires further evidence. The choice of supporting tools (\texttt{draw.io}, \texttt{WoPeD}, \texttt{ProM}, \texttt{Celonis}, \texttt{Signavio}, \texttt{Trisotech}, \texttt{n8n}, \texttt{Ollama}, \texttt{Qdrant}) is opinionated and may affect transferability.

	\section{Conclusion}
	\label{sec:conclusion}
	
	This paper presented the Banking case and the AI-decision checkpoint framework that structures it. Six checkpoints, one per BPM lifecycle phase, turn the lifecycle from a passive organizer into a chain of evidence-grounded decisions about AI adoption, instantiated in a Master-level course with a continuously evolving banking onboarding artifact.
	
	Formative evidence suggests three preliminary conclusions. First, checkpoints appeared to encourage students to treat AI as conditional on process-level value: groups revised initial over-estimates once simulation evidence accumulated, and incorporated governance structures in their BPMN and implementation artifacts. Second, the discovered process map from LMS's log characterizes submission-timing patterns and highlights two actionable adjustments: deferring cancellation patterns to M4 and adding a CP1 submission touchpoint for systematic data capture. Third, a voluntary group survey (10 of 24 groups) indicates that students perceived positive learning value across all modules (all medians~$\geq$~4), with M5 implementation activities receiving the highest ratings, consistent with the LMS evidence that prior checkpoints appeared to have effectively scaffolded that phase.
	
	For educators, the framework integrates AI into existing BPM courses without replacing the lifecycle backbone or mandating specific tools. The key transferable principle is the \emph{evolving-artifact} structure, which ensures students revisit AI decisions with accumulating evidence rather than treating each module as an independent exercise. Future work will extend LMS instrumentation for richer data capture per checkpoint and, after several course runs, codify recurrent reasoning patterns into structured checkpoint templates to support adoption in other institutional contexts.

	\bibliographystyle{unsrt}
	\bibliography{references}

\end{document}